\documentclass[english,round]{article}
\usepackage[T1]{fontenc}
\usepackage[latin9]{inputenc}
\usepackage{units}
\usepackage{textcomp}
\usepackage{amsthm}
\usepackage{amsmath}
\usepackage{amssymb}
\usepackage{graphicx}
\usepackage[authoryear]{natbib}

\makeatletter
\numberwithin{equation}{section}
\numberwithin{figure}{section}

\makeatother

\usepackage{babel}
\begin{document}

\title{Artificial general intelligence through recursive data compression
and grounded reasoning: a position paper}

\author{Arthur Franz%
\thanks{e-mail: franz@fias.uni-frankfurt.de%
}}

\date{January 6, 2015}
\maketitle
\begin{abstract}
This paper presents a tentative outline for the construction of an
artificial, generally intelligent system (AGI). It is argued that
building a general data compression algorithm solving all problems
up to a complexity threshold should be the main thrust of research.
A measure for partial progress in AGI is suggested. Although the details
are far from being clear, some general properties for a general compression
algorithm are fleshed out. Its inductive bias should be flexible and
adapt to the input data while constantly searching for a simple, orthogonal
and complete set of hypotheses explaining the data. It should recursively
reduce the size of its representations thereby compressing the data
increasingly at every iteration.

Based on that fundamental ability, a grounded reasoning system is
proposed. It is argued how grounding and flexible feature bases made
of hypotheses allow for resourceful thinking. While the simulation
of representation contents on the mental stage accounts for much of
the power of propositional logic, compression leads to simple sets
of hypotheses that allow the detection and verification of universally
quantified statements.

Together, it is highlighted how general compression and grounded reasoning
could account for the birth and growth of first concepts about the
world and the commonsense reasoning about them.
\end{abstract}
\pagebreak{}

\setcounter{tocdepth}{2}

\tableofcontents{}

\pagebreak{}

\section*{Introduction}

\addcontentsline{toc}{section}{Introduction}

This position paper contains a collection of ideas that I have developed
over the last years concerning the creation of a system exhibiting
artificial general intelligence (AGI). Although I came up with them
on my own, most if not all are not new and spread all over the literature.

The notion ``general'' is to be emphasized here. Unfortunately,
after early unsuccessful attempts research in artificial intelligence
(AI) has moved its focus on solving narrowly defined problems and
tasks, which became known as ``narrow AI'' \citep{kurzweil2005singularity}:
world level in chess, jeopardy, backgammon, self-driving cars, talking
personal assistants and a myriad of other commercial applications.
Although those are impressive achievements and the usefulness of such
applications is beyond any doubt, a system that exhibits general intelligence
seems still to be far away.

After compiling a large set of definitions in the literature \citet{Hutter:07idefs}
came up with a definition of intelligence that is consistent with
most other attempts:

``Intelligence measures an agent's ability to achieve goals in a
wide range of environments.''

This is exactly, what narrow AI does not achieve: it is programmed
for a very specific well-defined set of environments. Any deviation
from that narrow set most probably leads to failure of the system.

Conversely, humans are usually able to solve all sorts of tasks in
very diverse environments. Moreover, neuroscientific evidence teaches
us, that brains are able to process data cross-modally, e.g.\ by
transforming visual data to auditory or tactile stimuli in sensory
substitution devices. It is also known that in newborn ferrets neurons
in the auditory cortex adopt characteristics of visual cells, if fed
with stimuli from the visual pathway \citep{sur1988experimentally}.
Those observations point to the hypothesis that the human brain is
a general processor of quite diversely structured data.

This idea is, of course, not new and is around at least since Simon
and Newell's General Problem Solver developed in 1957. Although the
problem is far from being solved practically, \citet{Hutter:04uaibook}
has developed a mathematical formulation and theoretical solution
to the universal AGI problem, called AIXI. Even though AIXI is incomputable,
a lot can be learned from the formulation and general thrust of research.
The basic idea is the following. An AGI agent receives input data
from its sensors and picks an action at every time step while trying
to maximized reward. All data can be expressed as a binary sequence.
In order to act successfully, sequences have to be predicted, which
is achieved through Solomonoff's universal theory of induction. Solomonoff
derived an optimal way of predicting future data, given previous observations,
provided the data is sampled from a computable probability distribution.
In a nutshell, Hutter defines AIXI by espousing the Bellman equation
of reinforcement learning to Solomonoff's sequence prediction.

\citet{solomonoff1964formal,solomonoff1978complexity} has defined
his famous universal prior that assigns a prior probability (a semimeasure
to be precise) to every sequence,
\[
M(x)\equiv\sum_{p:U(p)=x}2^{-\text{|}p|}
\]
where the sum is over all halting programs $p$ of length $|p|$ for
which the universal prefix Turing machine $U$ outputs the sequence
$x$. The universal prior exhibits an Occam bias: by far the most
probability mass is captured by short explanations (programs) for
an observation $x$. Impressively, Solomonoff has proved that this
prior correctly predicts any computable sequence: $M(x_{t}|x_{1},\ldots,x_{t-1})\rightarrow1$
as $t\rightarrow\infty$, where $x_{i}$ denotes the $i$th sequence
entry. In essence, we learn that if we are able to find short programs
for arbitrary sequences the problem of universal inference is provably
solved. Intuitively, the scientific method itself is about the search
of simple (short) explanations of phenomena. Arguably, it is a formal
and institutionalized reasoning method, but people, even infants,
seem use it in more simple everyday situations \citep{gopnik1999scientist}.
If understanding the world means to compress sensory data, we need
a general data compressor.

Unfortunately, Solomonoff induction is not computable. Therefore,
Hutter and colleagues have developed approximations to AIXI, e.g.\
a Monte-Carlo approximation that uses prediction suffix trees that
enable predicting binary D-order Markov sequences \citep{Hutter:11aixictwx}.
This is an impressive achievement leading to a single system being
able to play various games (Pac-Man, Kuhn poker, TicTacToe, biased
rock-paper-scissors, 1d-maze, cheese maze, tiger and extended tiger)
without having specifically been programmed for them -- a notable
step towards generality of AI. In spite of that, it seems questionable
whether this approximation can be extended any further beyond Markov
sequences, since a well-known computational problem awaits: the curse
of dimensionality. We will come back to that later.

It may seem not intuitive that data compression plus reinforcement
learning can lead to the solution of such diverse and non-trivial
tasks. Traditionally, one may suspect that various cognitive processes
must be involved in the solution of such tasks. Hutter shows how data
compression implicitly incorporates those processes. It may be objected
that simple deep search of a chess program also replaces all sorts
of reasoning processes that presumably go on inside a human chess
player's brain. However, AIXI is not a short-cut narrow-AI-like solution,
but provably a genuinely general approach. This has convinced me that
general data compression is the way to go if we head for general intelligence.

\section{Approaching general data compression}

\subsection{\label{sub:Simple-but-general}Simple but general}

\subsubsection{Simplicity of tasks}

\begin{figure}
\begin{centering}
\includegraphics[width=0.8\textwidth]{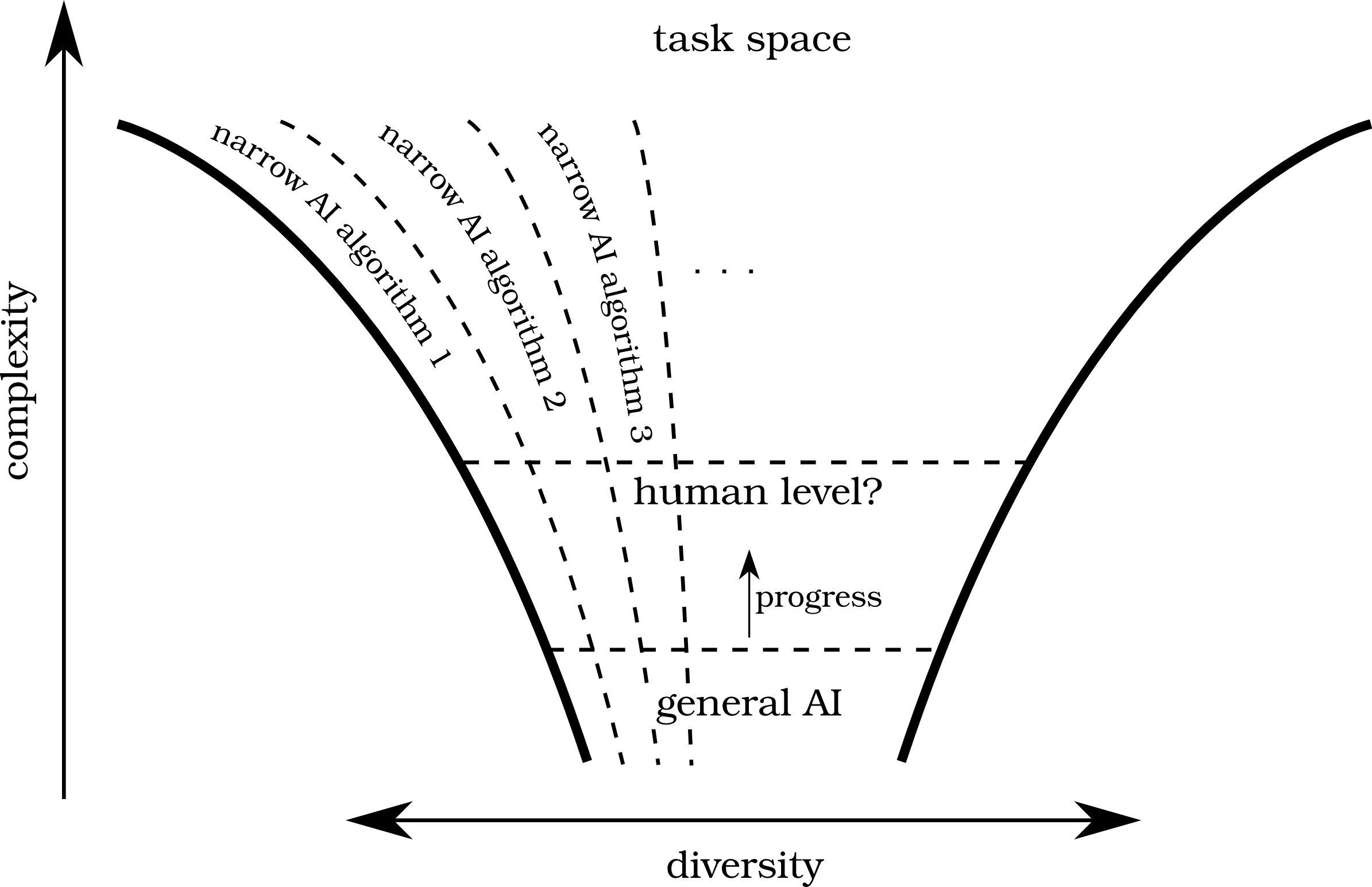}
\par\end{centering}

\caption{Approach to artificial general intelligence. Instead of trying to
solve complex but narrow tasks, AGI research should head for solving
all simple tasks and only then expand toward more complexity.\label{fig:agi-approach}}
\end{figure}

Given the form of the universal prior one may consider universal search.
For example, Levin search executes all possible programs, starting
with the shortest, until one of them generates the required sequence.
Although general, it is not surprising that it is a computationally
costly approach and rarely applicable in practice.

On the other side of the spectrum, we have non-general but computationally
tractable approaches: common AI algorithms and machine learning techniques.
Why could they not be generalized? The problem that all those techniques
face at some point is known as the curse of dimensionality. Considering
the (algorithmic) complexity and diversity of tasks solved by typical
today's algorithms, we observe that most if not all will be highly
specific and many will be able to solve quite complex tasks (Fig.\
\ref{fig:agi-approach}). Algorithms from the field of data compression
are no exception. For example, the celebrated Lempel-Ziv compression
algorithm (see e.g.\ \citealp{cover2012elements}) handles stationary
sequences but fails at compressing a simple non-stationary sequence
efficiently. AI algorithms undoubtedly exhibit some intelligence,
but when comparing them to humans, a striking difference comes to
mind: the tasks solvable by humans seem to be much less complex albeit
very diverse. After all, it is very hard for humans to perform depth
search in chess 10 moves ahead or learn the transition probabilities
of a variable-order stochastic Markov process, while they can do both
to some extent. For example, fitting the latter is performed by Hutter's
Monte-Carlo AIXI approximation. Although Hutter has found a general,
but incomputable solution to the AGI problem, in the Monte-Carlo approximation
he uses again a narrow-AI-like approach. Others try to fill the task
space by ``gluing together'' various narrow algorithms that would,
hopefully, synergistically cancel each other's combinatorial explosions
\citep{goertzel2009opencogprime}.

In a nutshell, I suggest that we should not try to beat the curse
of dimensionality mercilessly awaiting us at high complexities, but
instead head for general algorithms at low complexity levels and fill
the task cup from the bottom up.

\subsubsection{Simplicity of the algorithm}

Given that I have set the goal to compress general but simple data
sets, the question arises whether the algorithm that performs that
task can expected to be complex or rather simple as well. From the
point of view of ``narrow AI'' the programmer has to anticipate
exhaustively all data situations that his algorithm could possibly
be exposed to, which would otherwise lead to bugs. Such an approach
naturally leads to very complex and still not general algorithms.
However, as mentioned earlier, it is the very hallmark of generality
that the algorithm itself is required to be able to deal with the
whole variability of data situations. Does it mean that the general
AI algorithm could actually be quite simple itself?

A biological argument points in that direction \citep{berglas2008artificial}.
Human intelligence must ultimately be encoded in the DNA. The human
DNA consists of only 3 billion base pairs. Since there are four bases
(A, C, T and G), one base carries the information of 2 bits. Therefore,
the amount of information encoded in the DNA is merely $3\cdot10^{9}\cdot2/8/1024^{2}=715$
megabytes. It fits on a single Compact Disk and is much smaller than
substantial pieces of non-intelligent software such as Microsoft Vista,
Office, or the Oracle database.

``Further'', Berglas writes, ``only about $1.5$\% of the DNA actually
encodes genes {[}although it is currently debated whether the rest
is just redundant repetitive junk{]}. Of the gene producing portions
of DNA, only a small proportion appears to have anything to do with
intelligence (say 10\%). The difference between Chimpanzee DNA and
man is only about 1\% of gene encoding regions, 5\% non-gene. Much
of this can be attributed to non-intelligent related issues such as
the quickly changing immune system and human's very weak sense of
smell. So the difference in the ``software'' between humans and
chimpanzees might be as little as $715\cdot10\%\cdot1.5\%\cdot1\%=$11
kilobytes of real data.'' Of course, we are dealing with a quite
compact representation and Berglas may be wrong about one or two orders
of magnitude, but hardly more. ``In computer software terms even
$1.0$ megabytes is tiny.''

I therefore conclude that the algorithm for general intelligence,
at least as general as human intelligence, must be simple compared
to modern software. We are facing a software problem, not a memory
problem.

\subsection{\label{sub:partial-progress}A measure for partial progress in AGI}

One of the troubles of AGI research is the lack of a measure for partial
progress. While the Turing test is widely accepted as a test for general
intelligence, it is only able to give an all or none signal. In spite
of all attempts, we did not yet have a way to tell whether we are
half way or 10\% through towards general intelligence.

The reason for that disorientation is the fact that every algorithm
that achieved part of what we may call intelligent behavior, has failed
to generalize to a wider range of behaviors. Therefore, we could not
tell whether we have made some progress in the right direction or
whether we have been on the wrong track all along. As \citet{dreyfus1992computers}
cynically remarks, progress in AI is like the man who tries to get
to the moon by climbing a tree: ``one can report steady progress,
all the way to the top of the tree''. Since dead ends have been ubiquitous
there has been growing skepticism in the AI community.

However, since \citet{Hutter:04uaibook} has mathematically solved
the AGI problem (!), and the core part to be made tractable is the
compression part, we can formalize partial progress toward AGI as
the extent to which general compression has been achieved.

As I argued in ch.\ \ref{sub:Simple-but-general}, if we start out
with a \textit{provably general} algorithm that works up to a complexity
level, thereby solving all simple compression problems, the objection
about its possible non-generalizability is countered. The measure
for partial progress then simply becomes the complexity level up to
which the algorithm can solve all problems. Here, I will try to formalize
that measure.

Suppose, we run binary programs on a universal prefix Turing machine
$U$. $U$'s possible input programs $p_{i}$ can be ordered in a
length-increasing lexicographic way: ``'' (empty program), ``0'',
``1'', ``00'', ``01'', ``10'', ``11'', ``000'', etc. up
to a maximal complexity level $L$. We run all those programs until
they halt or for a maximum of $t$ time steps and read off their outputs
$x_{i}$ on the output tape. In contrast to Kolmogorov complexity%
\footnote{The Kolmogorov complexity of a string is defined as the length of
the shortest program able to generate that string on a Turing machine.%
}, we use the time-bounded version -- the Levin complexity -- which
is computable and includes a penalty term on computation time \citep{li2009introduction}:
\[
Kt(x)=\underset{p}{\min}\{|p|+\log t:U(p)=x\mbox{ in }t\mbox{ steps}\}
\]
Saving all the generated strings paired with their optimal programs
$(x_{i},p_{i}^{o})$ with $p_{i}^{o}=\{p:Kt(x_{i})=|p|+\log t,|p|\leq L\}$,
we have all we need for the progress measure. The goal of the general
compressor is to find all such optimal programs $p_{i}^{o}$ for each
of the $x_{i}$. If $p$ is the actual program found by the compressor,
its performance can be measured by
\[
r_{i}(L)=\frac{|x_{i}|-|p|}{|x_{i}|-|p_{i}^{o}|}\in[0,1]
\]
if the current string $x$ is among the $\{x_{i}\}$. If not, there
is no time-bounded solution to the compression problem. The overall
performance $R$ at complexity level $L$ could be used as a measure
for partial progress in general compression and be given by averaging:
$R(L)=\left\langle r_{i}(L)\right\rangle $. For example, one could
start with a small $L$ until $R$ approaches $1$ and increase $L$
gradually as suggested by Fig.\ \ref{fig:agi-approach}. 

One may object that the number of programs increases exponentially
with their length such that an enumeration quickly becomes intractable.
This is a weighty argument if the task is universal search -- a general
procedure for inversion problems. However, I suggest this procedure
to play the mere role of a test case for an \textit{efficient }general
compression algorithm, which will use completely different methods
than universal search and the properties of which shall be outlined
in ch.\ \ref{sec:Properties-of-the-gp}. Therefore, using the set
of simple programs as a test case may be enough to set the general
compression algorithm on the right track. If the limit complexity
of what is tractable today is $L_{\mbox{today}}$, then I doubt that
there exists an algorithm that is able to compress \textit{all} sequences
$x$ that can be generated by a program $p$ with $|p|\leq L_{\mbox{today}}$.
It will be a matter of future tests to find out.

\subsection{How many sequences are constructively compressible?}

It is well known in the theory of Kolmogorov complexity that most
strings cannot be compressed; more precisely, only exponentially few
$O(2^{n-m})$ binary strings of length $n$ can be compressed by $m$
bits (see e.g.\ \citealp{sipser2012introduction}). Interestingly,
the number of predictable sequences are also tightly bounded by an
expression of the same order of magnitude $\Theta(2^{n-m})$ \citep{kalnishkan2003many}.
This proven fact strengthens the intuition that understanding and
therefore predicting the world is about compressing sensory data.

Since we can not compress most sequences, it suffices to find programs
for that small fraction of compressible sequences. Further, we have
to be aware that an optimal algorithm that compresses all compressible
sequences may not exist. After all, it is not clear whether all information
needed to infer the shortest program is present in the sequence itself
or whether additional knowledge is required. For example, the first
few digits of $\pi$ ($3,1,4,1,5,9,2,6,5,3,\ldots$) may not contain
enough information to infer $\pi$, they rather follow only after
the discovery of additional knowledge about trigonometric functions
and their properties.

In summary, we are heading for an algorithm that infers short programs
generating most compressible sequences. The algorithm should be general
from the start, i.e.\ be able to find most if not all sequences below
a complexity threshold.

\section{\label{sec:Properties-of-the-gp}Properties of the general compressor}

Conventionally, when designing an algorithm, one is implicitly forced
to make a choice: either the algorithm is endowed with a strong inductive
bias towards a specific narrow class of data (e.g.\ linear regression),
which requires careful preparation of data and checking the requirements
of the algorithm (e.g.\ normality of distributions), or one uses
structures that can process broad classes of data, such as neural
networks, but leads to the curse of dimensionality. The former leads
to efficient inference but breaks down if the data is not in the appropriate
format. The latter is widely applicable but the struggle is with low
convergence rates, local minima or overfitting. In both cases careful
tuning is required by the programmer. Christoph von der Malsburg%
\footnote{Personal communication%
} diagnosed this situation quite cynically by saying that most of the
final algorithm's intelligence resides not in the algorithm itself
but in the programmer's intelligent tuning.

\subsection{\label{sub:Data-dependent-search-space}Data-dependent search space
expansion}

How shall we solve that dilemma? My suggestion is that the inductive
bias should change dynamically as data arrives.

To illustrate the idea, consider the following sequence:

$1,3,1,3,2,4,2,4,2,3,5,3,5,3,5,4,6,4,6,4,6,4,$

$1,4,1,4,2,5,2,5,2,3,6,3,6,3,6,4,7,4,7,4,7,4,...$

The first 4 digits may indicate that the sequence alternates between
1 and 3. This hypothesis is then expanded as a new alternation is
discovered subsequently between 2 and 4. This may lead to the hypothesis
that we are dealing with blocks of alternation subsequences. The next
block alternates 3 and 5 and we discover that each block is longer
by 1 element than the previous block, while the starting number is
also increasing from 1 to 2 to 3 and so on, while the difference between
alternating numbers is always 2. This hypothesis is changed again
when 1 and 4 start to alternate, hence it looks like the starting
number and block length has been reset and the difference is increased
to 3.

In fact a quite simple program can be written to generate that sequence.
But how could it be inferred? Humans obviously can do this. We notice
that the inductive bias and the corresponding search space is increasingly
expanded in directions dictated by the data itself. First, alternation
can be parametrized by two numbers -- a small search space quickly
instantiated with 1 and 3. Then it is expanded to represent blocks
of subsequences containing alternating sequences. Then, not two numbers
are saved, but the starting one and the difference (equal to 2 then
to 3) are saved. And finally the simplest parametrized representation
that contains the present sequence is found: blocks of alternating
sequences of variable differences and block lengths.

Conventionally, one would either preprogram this parametrization and
learning would simply consist of finding the parameters. Or one would
define a large search space of programs containing the correct one
and end up being lost in the search space. In contrast to that, I
suggest starting with a small search space and expand it in directions
imposed by the actual data.

Of course, I am not the only one who thought about this problem. An
interesting piece of work comes from the Bayes community. \citet{kemp2008discovery}
present an algorithm using hierarchical Bayesian inference in order
to ``discover structural form'', e.g.\ given feature vectors of
animal properties inferring that they should be arranged on a (evolutionary)
tree rather than on a chain, circle or grid. The interesting property
is that learning is reasonably fast given a quite large search space.
After all, the structural forms are not given a priori. Thus, an interplay
between several Bayesian hierarchies happens. First, a piece of data
comes in and produces a slight bias towards one of the structures.
Then, this slight inductive bias toward some of the structures is
used to categorize new data more efficiently, which leads to an even
faster formation of bias toward a structure. Hence, we see here a
nice example of a flexible inductive bias. The downside is though
that the overall large space of structures has to be defined, in this
case by a graph grammar generating the structures. Consequently, the
whole big search space is still given a priori, learning is mere selection
of one of the hypotheses in the large space; just inference is made
in clever way. In contrast to that, I suggest to refrain from defining
the search space of the algorithm before data arrives.

Actually, this insight should be obvious. After all, the scientific
method does not prespecify all possible theories that could explain
all possible worlds before starting to observe the world experimentally.
Instead, when new data comes in, scientists try to find a set of simple
explanations consistent with it and all previous data. In computer
science terms, a large search space is traversed efficiently by ruling
out large subspaces inconsistent with data. Solutions of considerable
complexity can be found that way, just think of modern theories in
physics.

Our line of reasoning suggests to the following iterative approach.
\begin{enumerate}
\item Look at a sufficiently small piece of data.
\item Construct a set of as simple as possible hypotheses consistent with
it -- a small search space.
\item Look at the next piece of data and compute the likelihoods and posteriors
of the hypotheses.
\item Expand the search space around the most likely hypotheses, e.g.\
find generalizations or supersets of the most likely hypotheses. Discard
the unlikely ones.
\item Go to 3 until the posterior probability of a hypothesis is large enough.
\end{enumerate}

\subsection{\label{sub:Features-and-hypothesis}Features and hypothesis sequences}

The crucial question becomes how to construct the set of simplest
hypotheses consistent with the sequence part seen so far. If we solve
this problem for arbitrary sequences, I suspect that the most difficult
task for a general data compressor will be solved.

For example, consider a sequence, starting with $1,3,\ldots$ Suppose
one considers the null hypothesis that it is a deterministic first-order
Markov sequence using addition. Then, the only unknown is the summand
which can be fitted to be $2$ -- a small search space --, and the
sequence can be continued to $1,3,5,7,9,\ldots$ There are several
ways the null hypothesis can be expanded: it can be \textit{questioned}
in three possible ways. The sequence could be
\begin{itemize}
\item indeterministic,
\item higher-order Markov or non-Markovian at all, or
\item using a different arithmetic function or an arbitrary function,
\end{itemize}
or a combination of any of them. It seems to be a general observation
that the definition of a hypothesis consists of features (here determinism,
Markovianity and the applied function) that can be questioned systematically
and from which new hypotheses can be derived. The null hypothesis
is a specification of the search space, the inductive bias, within
which a search algorithm has to find a solution. The ``narrow AI''
approach is marked by the fact that such specifications are provided
by a human programmer after a careful prior examination of the data
set. Only the remaining ``blind'' search is performed by the algorithm,
which is then proudly announced to be ``intelligent'' \citep{mcdermott1976artificial}.
If we want to depart from such practices, we have to find an algorithmic
way to question those specifications and corresponding underlying
assumptions.

For example, a higher order Markov process can be described by taking
data from $n$ previous entries implying position offsets described
by the family of sets $\{-1,-2,\ldots,-n\}$ parametrized by $n$.
Searching for a solution in this subspace is what I call \textit{expanding
the hypothesis in the direction of the feature}. Expanding in the
direction of Markovianity thus leads to a sequence of possible alternative
hypotheses, ordered after complexity: 2nd, 3rd,... , $n$-th order
Markov processes. This ordered set of alternative hypotheses in the
direction of a particular feature is what I call \textit{hypothesis
sequence}.

Overall, it seems that features act as a ``basis'' and elements
of a hypothesis sequence act like ``coordinates''. Specifying the
value of each feature leads to a sufficient specification of the problem
for a search algorithm to solve it. The metaphor of a feature basis
will prove useful, as we will see later, and hopefully could move
beyond a metaphor and acquire a precise mathematical meaning at some
point.

\subsection{Measuring progress: the compression rate}

Compressing data means finding a representation of it that takes less
memory. In our case, we want to infer programs that generate a sequence.
Consider the finite string with length 16: $0001111100000000$. Assuming
it to be defined on the domain $\{0,1\}$ and each entry drawn from
it with probability $p=0.5$, then its entropy is $H_{0}=-16\ \log_{2}(p)=16$
bits. It takes 16 bits of memory to store it. Suppose, we have inferred
a parametric program that represents ``start at position $n$ and
write $l$ ones, all others are zero''. As $n$ and $l$ can range
between 1 and 16, each of them requires $H_{\mbox{pars}}=\log_{2}16=4$
bits to be specified. Additionally the program itself requires memory
$H_{\mbox{prog}}$. The goal is to maximize the compression rate
\[
1-\frac{H_{\mbox{prog}}+H_{\mbox{pars}}}{H_{0}}
\]

each time a new representation is found.

\subsection{Recursiveness}

Suppose, such blocks of ones occur 10 times in a string of length
1024. Then specifying all starting positions and lengths takes $H_{\mbox{pars}}^{(1)}=2\cdot10\cdot\log_{2}(1024)=200$
bits. Neglecting the size of the program it corresponds to a considerable
compression rate of $1-200/1024=80.4\%$. But suppose we discover
a regularity in the starting positions and lengths, say $n_{i}=100\cdot i$
and $l_{i}=4$. Then only one length has to be specified and the step
size ($100$), which takes only $H_{\mbox{pars}}^{(2)}=20$ bits and
pushes the overall compression rate to $1-20/1024=98\%$. In this
fashion, data can be compressed recursively in the sense that the
same data compression machinery is first applied to the data itself
and then recursively to the parameters of the models. Here we notice
the need for the generality of the data compressor: after all, we
had a single binary sequence at first and then two integer sequences
for the starting positions and lengths, respectively. A recursion
level should be accepted if compression is increased. Recursive data
compression can reach arbitrary high levels until no more compressive
model is found. A great example in science is the quest for unification
in physics: the standard model of particle physics is left with only
19 parameters to be explained in a grand unified theory. In contrast
to this, AI algorithms usually do not possess additional compression
levels, except in the small field of metacognition research \citep{cox2005field}.
But even there compression is not recursive, i.e.\ different algorithms
are used at meta-levels, except some notable examples from Marvin
Minsky's group \citep{singh2005one,morgan2013substrate}. If we want
to build a general compressor though, there is no way we can foresee
which type of algorithm is needed for which data set and at which
level: the compressor has to be general enough to handle them all.

\subsection{Orthogonality of the feature basis}

In ch.\ \ref{sub:Features-and-hypothesis} we gave an example of
three features. Those features are \textit{orthogonal }in the sense
that the specification of each of them does not contain information
about any of the others. Markovianity does not bear on determinism
of the applied function, nor does (in)determinism specify dependence
structures or the applied function etc. Formally, the pairwise mutual
information between all features should be zero. We should aim to
find an orthogonal feature basis for the description of a data set
since otherwise features share information and lead to redundancy
in the representations, which implies a lower total compression rate.
Orthogonality also specifies the search procedure. Suppose we have
answered the question about the dependence structure and found out
that the current sequence entry only depends on the previous one.
Then all remaining questions \textit{boil down} to describing that
dependence and can be tackled independently. In other words, only
orthogonal features need to be considered.

In summary, our compression algorithm can be characterized as follows.
First, it has to find an orthogonal feature basis and expand in the
direction of those features. Then the likelihood of each hypothesis
in the space spanned by the feature basis can be computed. Then the
hypothesis with the largest posterior probability can be expanded
further etc.\ while we look at more and more data. We should use
the posterior instead of the likelihood since the Bayes theorem automatically
takes care of Occam's razor -- the trade off between explanatory power
of a model and its complexity (see chapter ``Model comparison and
Occam's Razor'' in \citealp{mackay2003information}). In parallel,
since the feature basis is parametrized, the residual entropy in the
parameters should be compressed further in higher recursion levels
leading to increasingly simpler and powerful models.

\subsection{\label{sub:Extracting-orthogonal-features}Extracting orthogonal
features}

After establishing orthogonality the search space for a feature basis
is severely reduced. Nevertheless, features have to be extracted somehow.
In Principal Component Analysis (PCA) at each step a vector is first
defined pointing to the direction of largest variance and then the
data cloud is projected onto the surface perpendicular to that vector
such that the variance in its direction is nullified. Subsequently,
only the residual variance in perpendicular directions is considered
such that ultimately an orthogonal basis is found ordered after the
variance ``explained'' by the vectors. Analogously, orthogonal features
can be extracted when focusing on the residual variance of the data.
For example, consider a point B lying exactly in the middle between
two other points, A and C. Suppose the features ``distance'' and
``angle'' are available to the system. First, the system would notice
that the distances A-B and B-C are equal and thereby discover the
equidistancy feature, since two equal distances leads to compression
(see ch.\ \ref{sub:Conceptualizing-objects}). This feature may be
viewed as the first ``principal component''. Then, images can be
\textit{sampled} holding the equidistancy feature active, which results
in random isosceles triangles. Subsequently, the residual variance
is found in the angle feature which is 180\textdegree{} (or $\pi$)
-- again a compressible number -- in the case of B lying in the middle
between A and C. This leads to the discovery of the ``between''
feature. In this way, the situation ``B is in the middle between
A and C'' can be described in a complete and orthogonal feature basis:
middle = equidistant and between. The basis is orthogonal, because
angles and distances can be changed independently of each other. It
is complete, because constraining a point to be in equal distance
to two other points while lying at the same time between them, necessarily
produces instances of the ``middle'' situation.

Even though the requirement of an orthogonal and complete basis and
a PCA-like procedure for its search greatly reduces the search space
for features, it is not clear enough to me how to find features in
arbitrary data situations and constitutes one of the frontiers for
future research. A crucial, feature defining step seems to be the
ability to realize that the current data situation is a special case
of a general one. After all, as I argued in ch.\ \ref{sub:Data-dependent-search-space},
the general description must not (and can not) be given a priori.

\subsection{\label{sub:Interpretation-rivalry}Interpretation rivalry}

Someone said that the idea of splitting the AI field in a multitude
of subfields has marked the beginning of the failure of the whole
field. After all, perception requires reasoning, reasoning requires
learning, learning has to rely on planning an vice versa -- all subfields
are actually densely interconnected in the human mind. Hence, the
attempt to solve them separately from the others may have slowed down
the progress in \textit{general} AI.

For example, image segmentation in computer vision suffers from the
problem that our ability to segment an image into separate objects
and their parts heavily depends on our knowledge about the objects.
As I will argue in ch.\ \ref{sub:Conceptualizing-objects}, the conceptualization
of objects is driven by compression. Therefore, an image or sequence,
should be segmented in such a way that compression is maximized. I
suggest that a sequence should be segmented in those cases when all
segments are highly compressible while the whole sequence can not
be easily compressed. For example, piecewise constant sequences: 15,
15, 15, 15, 32, 32, 32, 32, 32, 32, 7, 7, 7, 7. Implicitly, it was
assumed here, that the segmentation consists in finding a partition
of the position set $\{1,\ldots,14\}$ into intervals, and not an
arbitrary partition. This bias can again be explained by the recursiveness
of general compression: intervals can be described simply by two numbers
while the description of a generic subset would have to enumerate
all positions that it consists of.

In essence, the problem consists of finding an assignment of every
data point to a subset of the partition. Like in multistable perception
images where the same image can be interpreted in several ways, one
can frame the problem as a rivalry for different interpretations of
the same image, while trying to maximize compression. The problem
is reminiscent of the famous Ising model in which the spins of a hot
ferromagnet are first oriented randomly, but increasingly form islands
of equally oriented spins as the temperature decreases. This behavior
is explained by a high energy that is required to keep neighboring
spins in opposite directions. Similarly, a perceptual scene should
naturally break up into segments/objects when trying to maximize compression.

\section{Grounded reasoning}

Although general data compression seems to be a central ingredient
for AGI, several other important issues like language, memory, reasoning,
commonsense knowledge, resourcefulness, brittleness and many others
keep staring at the researcher intimidatingly. In the following, far
from claiming to have solved anything, I will introduce my ideas about
some of them and highlight the way, general data compression bears
on them.

Suppose, general compression works. What then? One of the most burning
questions in AI is the problem that AI systems do not really know
anything about the world, commonsense knowledge possessed by any 3-year-old.
As Marvin Minksy has put it, ``no program today can look around a
room and then identify the things that meet its eyes'' \citep{minsky2011interior}.
The problem is not just about identification but about being able
to understand and describe the objects and knowing about their function.

In this section, I will argue why grounded reasoning is an important
step towards commonsense reasoning an thereby towards AGI and how
it densely and naturally interacts with general compression.

\subsection{\label{sub:What-is-grounding}What is grounding?}

In his seminal paper, \citet{harnad1990symbol} addresses the so-called
symbol grounding problem -- a symptom of a disease of purely symbolic
AI systems:

``How can the semantic interpretation of a formal symbol system be
made intrinsic to the system, rather than just parasitic on the meanings
in our heads? How can the meanings of the meaningless symbol tokens,
manipulated solely on the basis of their (arbitrary) shapes, be grounded
in anything but other meaningless symbols?''

I suggest to split the problem into two subproblems. The first is
a philosophical problem called intentionality: how can symbols or
any representations for that matter re-present anything about the
world? Where is the invisible arrow pointing from the symbol DOG to
the real dog? And how can the symbol DOG ever express the ineffable
meaning of a real dog? Based on such questions, there is a huge philosophical
discussion about whether computers could ever think (see e.g.\ \citealp{dreyfus1992computers});
after all, computers only juggle the symbols ``0'' and ``1'' around,
without ever being able to know what anything truly means. I shall
not dive into this discussion, but merely state that representations
are not meant to possess any intentionality, there is no arrow, but
merely a mechanical reaction to external stimuli. The ``grandmother
neuron'' simply reacts to the occurrence of the grandmother in the
visual field, it does not point to the grandmother in any way, nor
does it ``know'' about the grandmother in any deeper epistemological
sense.

\begin{figure}
\begin{centering}
\includegraphics[width=0.6\textwidth]{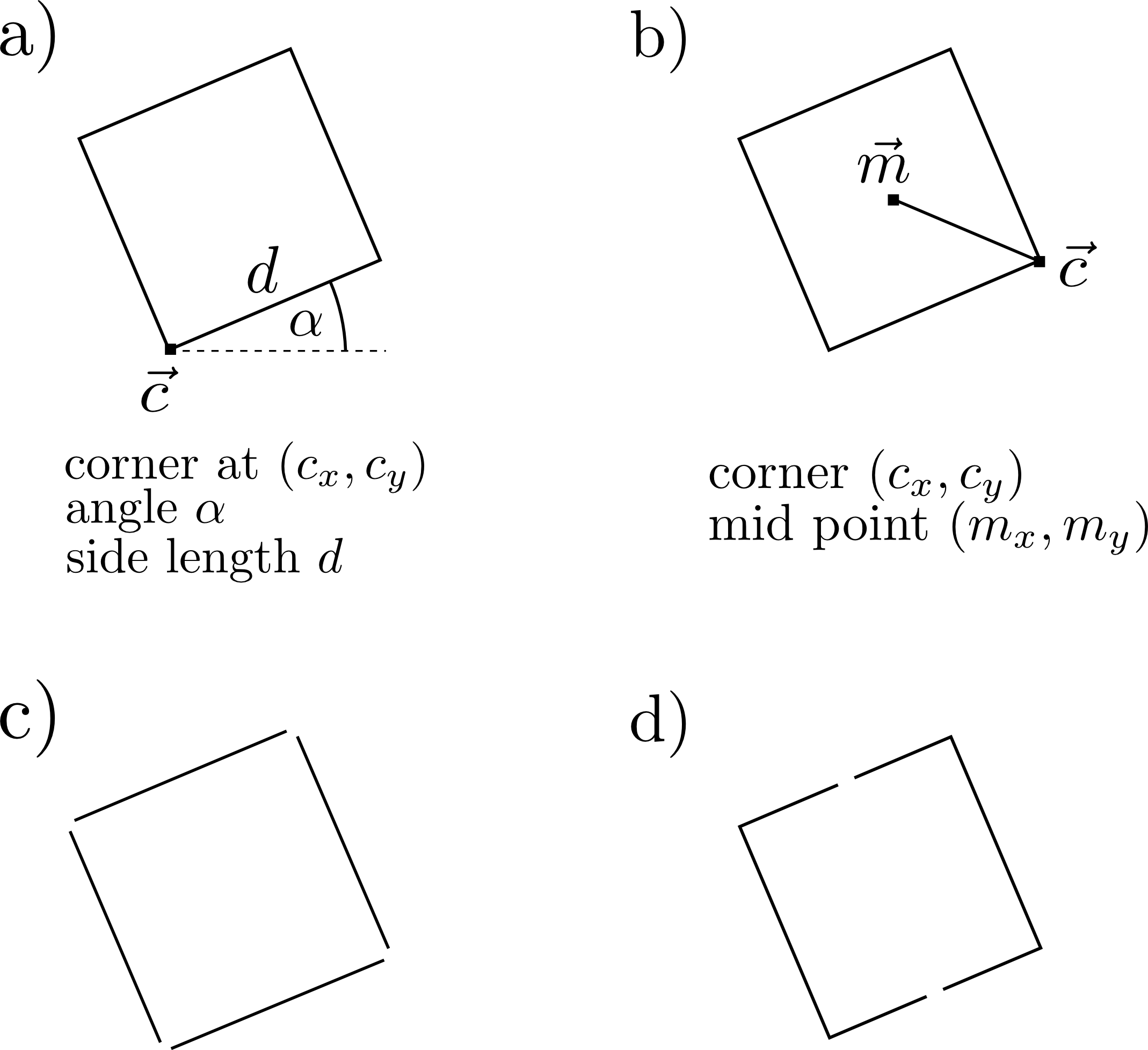}
\par\end{centering}

\caption{Possible feature bases representing a particular square.\label{fig:feature-bases-square}}
\end{figure}

The second subproblem is more important though. I define a system
as grounded if it is able to form representations at arbitrary granularity.
Imagine several feature bases of a square (Fig.\ \ref{fig:feature-bases-square}).
It is easy to see, that four features are enough to specify the square,
that is to form a complete, orthogonal basis for it. But the basis
is not unique, Fig.\ \ref{fig:feature-bases-square}a is just as
good as \ref{fig:feature-bases-square}b. Further, a basis can be
formed from line segments as in Figs.\ \ref{fig:feature-bases-square}c
and \ref{fig:feature-bases-square}d. However, it is the hallmark
of features to represent an aspect of the stimulus while dismissing
other information. Once the line feature is represented, it can only
be changed by its parameters (end point coordinates), but not cut
into pieces as in \ref{fig:feature-bases-square}d. One the feature
base in \ref{fig:feature-bases-square}a is chosen, changing $\alpha$
leads to a rotation around the corner, but not, say, around the mid
point of the square. The point is, once a feature basis of an object
is chosen, the representation becomes atomic, such that the ability
to form more fine-grained representations is lost. The only way to
split the atoms is to go back to the low level input, either to the
actual square stimulating the systems sensors or to generate an imagined
square from the complete feature basis. Only then a new, more fine-grained
feature basis can be extracted, since only the low level input stimulus,
presented to the system point by point, contains enough information
to accomplish that task.

It may be interjected that it is exactly those variably fine-grained
representations that hierarchical network approaches such as DeSTIN
\citep{arel2009destin} or Hawkins' Hierarchical Temporal Memories
\citep{hawkins2007intelligence} develop at each of their levels.
However, I see at least three problems associated with them. First,
nodes at each level look at a specific, hard-wired patch of child
nodes at the level below. Thus, a hierarchical segmentation of the
image is essentially hard-wired and fails to fulfill the requirement
of a data-dependent inductive bias (ch.\ \ref{sub:Data-dependent-search-space}).
Second, such hierarchies are designed for finding partonomic stimulus
decompositions, while failing to find other representations (e.g.\
taxonomies). Finally, the representations can not be transformed,
thus inhibiting resourceful thinking. The latter point is so important
that the next subsection will be devoted to it. Nevertheless, hierarchical
representations may well be the right solution, but their flexibility
has to increase significantly, probably to obtain the same expressive
power as the hierarchies of program trees in general.

Consider further the procedure in ch.\ \ref{sub:Extracting-orthogonal-features}.
Assume a function that establishes whether a point is in the middle
between two others has been hard-coded, making it impossible for the
system to analyze it further. However, remarkably, this PCA-like procedure
allows for the decomposition of a seemingly atomic/symbolic concept
``middle'' into its components ``equidistant'' and ``between''.
A grounded system is able to dissolve a concept such as a ``grandmother''
into its conceptual components (e.g.\ body parts), the components
recursively into their own components etc.\ down to the raw input
image. It is thus the interaction between the concept and the low
level input that allows for an analysis of variable granularity.

As we shall see, grounding will set the foundation for non-symbolic,
context dependent reasoning and resourceful thinking.

\subsection{Resourcefulness}

Intelligent problem solving requires the ability to think in different
ways about the problem, which Minsky coined as resourcefulness \citep{minsky2006emotion}.
The research in the phenomenon of ``insight'' in thought psychology
constitutes a well presentation of this issue. For example, in Karl
Duncker's famous ``candle problem'' the subject is asked to fix
a lit candle on a wall (a cork board) in a way so the candle wax won't
drip onto the table below. To do so, the subject is provided a book
of matches and a box of thumbtacks. Usually, subjects have difficulties
to solve the problem until it dawns on them that the box containing
the thumbtacks, can be tacked to the wall and serve as holder of the
candle. Hence, humans have got the ability to think of a box sometimes
as a container and sometimes as a supporting device.

In the so-called mutilated chessboard problem \citep{kaplan1990search},
the two diagonally opposing corners of the $8$ x $8$ board are cut
out. The subject is required to either cover this mutilated chessboard
with domino pieces, each covering two squares, such that all 62 squares
are fully covered, or to prove that the task is insoluble. Usually,
after trying various coverings subjects experience an ``aha'' moment
realizing that the two missing corners are of the same color, say
white. Therefore, there are two more black squares than white squares.
Since each domino piece covers one black and one white square, there
will always be two black squares left after each covering of the board.
And since two black squares are never adjacent on a chessboard, the
task is insoluble.

The reason for the present discussion is that in such cases subjects
need to switch to a different ``representation space'' \citep{kaplan1990search}
of the problem -- a different feature basis in our terminology --
in order to solve it. First, the search space is spanned by the combinations
of positions of domino pieces. Only after attending to the color feature
of the chessboard and spanning the search space by the colors and
numbers of squares, the problem can be solved efficiently. Only after
switching from the containing to the supporting feature of the thumbtack
box, the solution of the candle problem comes to mind.

As it was discussed in ch.\ \ref{sub:What-is-grounding}, a square
can be represented in different ways. A intelligent system has to
be able to both understand that fact and to switch between various
representations in order to be resourceful.

Resourcefulness poses by itself a strong argument against ``narrow
AI'': if the representation of the problem is chosen a priori there
is no way the system could change it. Whether the representation is
symbolic or subsymbolic in nature, it intrinsically introduces a fixed
induction bias to the system. As I have argued in ch.\ \ref{sub:Data-dependent-search-space},
in spite of being important and necessary the induction bias must
be changeable in a flexible way.

This realization begs the question whether it is possible to find
an algorithm for the transformation between features bases. Is there
a general way to detect appropriate transformations, rotations in
search space, so to speak? What connects two representations of the
same data? Can such changes in representation can generally be achieved
without going back to the input itself?

I suspect that the answer is negative; after all only the low level
input contains all the information necessary to construct arbitrary
representations. A plausible way to switch representations is to take
the current one, \textit{generate} a data sample from it (simulate
it on the ``mental stage'') and look for a different feature basis
to represent it again. For example, one should take a definition (a
generative model) of a square, sample a particular square from it,
observe some other features of it and span a different feature basis,
which constitutes a different definition of the same figure. Without
the input, there is no way to \textquotedblleft{}cut through\textquotedblright{}
the existing symbols; neither transformations into different feature
bases seem possible without a severe task-specific formalization effort,
nor are representations of variable granularity possible.

Therefore, resourceful thinking is only possible if any construction
of the system's symbols must be performed \textit{via} the input.
Otherwise, it is either not possible to transform representations
or -- as it happens in formal logic -- transformations decouple the
symbols from the world/input, leaving them ungrounded, \textquotedblleft{}dangling
in the air\textquotedblright{} and independent of context.

In a nutshell, not only should the search for the right hypothesis
describing the data depend on the data itself, but also the resourcefulness
of thinking -- the ability to represent the data in different ways
-- should be tightly tied to the data itself. As such, resourcefulness
constitutes another argument for grounding.

\subsection{\label{sub:Formal-logic-for}Formal logic for commonsense?}

The AI community has been aware of the commonsense problem for quite
a while, but tackled it with limited success, unfortunately \citep{mueller2010commonsense}.
I suggest that the main reason for it is the lack of grounding of
representations.

The grounding of representations has been neglected for quite a while
although there have been calls for it \citep{harnad1990symbol,barsalou1999perceptions}.
Originally, the call was for grounding of symbols, since it is with
symbols that reasoning has been represented, through usage of formal
logic mostly.

Consider a commonsense problem, such using a string to tie a plant
to a rod, that is stuck into the ground. The prevalent method in the
commonsense reasoning community is to formalize the problem, such
that all sorts of valid statements can be logically concluded from
the formal logical system. The idea is to hand code abstracted relations
between entities of the situation, add some arguably general properties
of space and time, and then being able to express all other relationships
in the situation basically through combinations of those abstractions,
e.g.\ by forward chaining through the formal knowledge base.

There are several problems with such an approach. First, in practice,
commonsense situations are very difficult to formalize and arguably
the formalization process has to be done for each situation separately
unless one has formalized the whole world somehow. Of course, this
is exactly the ambition of projects like Cyc which try to do that
since 1985 \citep{lenat1995cyc}. Difficulty is not a formal argument
of course, but we have to ask ourselves whether we are on the right
path with such an approach.

Second, being ungrounded means having to carry the structure of a
real world situation into one's system which bears a practical danger
that the hand coded or concluded relations between entities will not
hold as the system is scaled up \citep{joscha2009principles}. There
is no guarantee that the encoded relations are actually the correct
abstractions from the world situations.

Third, ungrounded representations have difficulties reacting to new
situations where a new context leads to different conclusions.

Therefore, I suggest a simulation theory approach, much along the
lines of \citet{barsalou1999perceptions}, that keeps a tight connection
to the world while being able to reason about it, as I will argue
in the subsequent chapters.

\subsection{The world as its own model: reasoning without formal logic}

Up to now, we have mostly talked about representations and how to
switch between them, but not how to reason about properties and relations
in a grounded way. 

Consider dropping a perpendicular from a corner of an isosceles triangle
onto the base. Then we will land exactly in the middle of the base.
That sort of task is quite easy to formalize and a formal proof that
this is true for all isosceles triangles can be derived. On the other
hand, consider the approach of a grounded mental simulation, which
is arguably the way, humans solve the task \citep{barsalou1999perceptions}.
From a representation of an isosceles triangle a \textit{particular}
sample triangle is drawn onto the mental stage. Then, the perpendicular
is dropped that \textit{happens} to land in the middle of the basis,
a fact that can be \textit{read off from the mental stage} using appropriate
features. The main difference between the ungrounded and the grounded
approach is that the former tries to arrive at conclusions through
proofs, that is by transforming one's own representations, while the
latter samples a particular situation top-down onto the mental stage,
imposes the appropriate conditions and then reads off the result though
bottom up activation of features. This is the kind of grounded reasoning
employed by simulation theories.

Note that this sampling procedure implicitly computes a sort of modus
ponens as in propositional logic. Given an isosceles triangle, it
follows that the perpendicular onto the base will split it in half.
Interestingly, only the premise -- the presentation of a particular
isosceles triangle in the simulation space/mental stage -- is provided
by the system. The implication clause itself is not represented anywhere
in the system and still the conclusion can be \textit{measured} from
the mental stage. However, the implication clause is necessary for
modus ponens to work. Where is it then?

It must be in the world itself then. I suspect that \textit{it is
the structure of the world that implicitly ``represents'' by far
the most knowledge}. Either by perceiving the world, or by connecting
with it by mental simulation, it seems possible to get access to that
knowledge. In essence, it is not by hand coding or self-organized
learning of large knowledge bases, but by letting the world itself
``represent'' the large part of our knowledge, the commonsense knowledge
problem could be solved. Although this claim parallels Dreyfus' \citeyearpar{dreyfus1992computers}
call for Heideggerian AI, I do not share his rejection of representations
per se: we do need representations in our systems, but they shall
be grounded in the world and the world's intrinsic structure should
be used for reasoning. Also, it is not the magic touch of reality
that is sometimes suspected behind the successes of today's fashionable
``embedded, embodied cognition'', but a call for a tight connection
between one's representations and the world. It is the informational
richness of an actual image -- be it real, virtual or imagined --
that performs that intrinsic reasoning task.

Without going into details, other elements of logical reasoning, conjunctions,
disjunctions, resolution etc.\ can be performed by simulation theories
\citep{uchida2012perceptual}.

It may be objected that it may be difficult to construct grounded
knowledge of abstract concepts, while they are seemingly easy to construct
in formal logic, e.g. loves(father, son). However, one shall not be
fooled by the meaning that those symbols convey to us, since they
do not ground the machine but just remind us of our own groundedness.
I suspect that symbols tend to merely seduce the researcher to look
for shortcuts around the grounding problem. Conversely, even though
simulation theories seem only to be describing reasoning with visualizable,
commonsense objects, one shall not underestimate the power of analogical
reasoning in the formation abstract concepts. After all, the whole
tradition of empiricist philosophy argues that the acquisition of
abstract concepts may ultimately be grounded in perception.

\subsection{\label{sub:Universal-quantification}Universal quantification}

An important problem to solve in simulation theories is universal
quantification. After checking truth values of statements for some
particular simulated samples, how can it be inferred that the same
truth value will be observed for all samples under the present conditions?

Consider a probabilistic account: what are the chances of picking
a random isosceles triangle such that the base \textit{happens} to
be cut exactly in the middle by the perpendicular? Not high, and the
probability will decrease even more when measuring the cutting point
with higher precision. Of course, this is not a strict proof, but
as has been argued many times, human intelligence does not have to
be perfect, but just good enough for correctly dealing with most situations
in life.

If the system maintains hypotheses about the triangle in the background
of reasoning, it can evaluate the likelihood that a particular unpredicted
curiosity of the sample could have been generated by chance. Of course,
one would have to specify what such a curiosity is and how it is to
be identified. After all, why is cutting the basis at length fraction
of $0.5$ (in the middle) so much more suspicious and curiosity awakening
than cutting it at $0.46878$?

Here is where compression plays a central role. Choosing a binary
representation, the number $5$ can be expressed by $101$, whereas
$46878$ will be much longer ($1011011100011110$). In terms of Kolmogorov
complexity, the shortest program with $0.5$ as output will be much
shorter than the shortest writing $0.46878$. 

Keep in mind that all that was used by the simulation was just the
definition of an isosceles triangle. Without proof, the fact that
the base was met in the middle is surprising since the cutting point
is at a position of low complexity. The surprise comes from the fact
that a line piece can only have very few such points of low complexity
(low fractions as $\nicefrac{1}{2},\nicefrac{1}{3},\nicefrac{1}{4}$
or some other salient numbers such as $1/\pi$) and hence the probability
of hitting those points is very low unless entailed by the definition
of the problem in the first place.

Consider another example. Imagine a ``whip'': one end of a string
is tied to an end of a rod. In what case the other end of the string
could reach the other end of the rod? Commonsense dictates that the
string be at least as long as the rod. How did we arrive at that hypothesis?
Suppose, after a trial and error phase, the system figures out that
the length $l$ of the string is important. What condition should
be imposed on it? The system could invent an array of numbers $l_{1},l_{2},l_{3},\ldots$
and construct an arbitrary complex condition from it, such as $l\leq l_{1}\wedge l>l_{2}\wedge l\leq(l_{3}-l_{2})^{2}/l_{1}$.
However, the Occam bias dictates parsimonious solutions. Since the
length of the rod $l_{r}$ is already present in the data and does
not have to be invented, it is to be used preferably and in a simple
way. One of the simplest ways that is consistent with previous trial
and error data is therefore $l\geq l_{r}$. This hypothesis can be
quickly tested by choosing $l=l_{r}\pm\epsilon$ with a small $\epsilon$,
which will drive its posterior probability close to $1$. Just as
in sequence prediction, the reuse of already present variables such
as previous entries or the rod length, and doing it in a simple way
maximizes the chances of finding correct hypotheses.

Note that a hypothesis could be found \textit{for all} isosceles triangles
and for all strings complying with the conditions, which establishes
that \textit{simple hypotheses are viable candidates for valid universally
quantified statements}. Nevertheless, how much certainty can be gained
that the statement is really universally true and not just for the
few examples?

Getting a few examples fully consistent with the simplest explanation
is so compelling in terms of posterior probability that we arrive
close to certainty, because a simple explanation is so much more probable
a priori than a complex one. However, exceptions can always occur.
For example, two natural numbers $a$ and $b$ picked uniformly from
$1$ to $1000$ will be different with probability $99.9\%$. For
some reason, the exception, $a=b$ is exactly the compressible case,
since then only one number has to be stored. I therefore suspect that
exceptions occur preferably at compressible instantiations of the
variables, which considerably simplifies their detection. Otherwise,
if Nature wants to introduce exceptions at incompressible locations,
she has to pay for it with information. After all, since exceptions
are nothing but missing truth conditions of a statement, they are
thus biased towards simplicity as any truth condition.

We conclude that general compression with its hypothesis sequences
ordered from low to high complexity has the potential to solve the
universal quantification problem in simulation theories, bringing
us much closer to the solution of the commonsense problem.

\subsection{Testing hypotheses, intervention}

Given a set of possible hypotheses, the important question about hypothesis
testing arises. As can be shown, structure learning in Bayesian networks
proceeds much faster, if it is possible to intervene and observe the
effects of the intervention \citep{pearl1988probabilistic}. Essentially,
the question is about setting up scientific experiments. Which actions
shall be chosen in order to gain most information about the data given
current hypotheses?

The solution is known as the principle of maximum entropy \citep{mackay2003information}.
Given a set of hypotheses and their prior probabilities coming from
both previous data and the Occam bias, the probability of every result
of an experiment can be computed. The maximum entropy principle states
that the action should be chosen in such a way that the entropy of
the probability distribution of the possible results is maximal. In
other words the all results should be expected to be seen with the
same probability given current hypotheses.

For example, in the previous chapter, the goal was to test whether
the other end of the rod of length $l_{r}$ is reachable by a string
of length $l$. We assume that the possibility of reaching that end
is described by $l\geq l_{0}$, and some lengths $l$ have already
been tested reducing the possible range of $l_{0}$to $a\leq l_{0}\leq b$.
The result of the task shall be given by the variable $X$, with $X=1$
meaning that the task is possible. Suppose, there are two hypotheses,
$H_{1}:l_{0}=l_{r}$ and $H_{2}:l_{0}$ is uniform. If $H_{1}$ is
true, then any string longer than the rod will lead to success, hence
the likelihood of $l$ is $p(X=1|l,H_{1})=\Theta(l-l_{r})$, with
$\Theta$ being the Heaviside step function. If $H_{2}$ is true,
then the probability of success increases linearly between $a$ and
$b$, $p(X=1|l,H_{2})=\frac{l-a}{b-a}$. Marginalizing out the hypotheses,
we get
\[
p(X=1|l)=p(X=1|l,H_{1})p(H_{1})+p(X=1|l,H_{2})p(H_{2})=
\]
\[
\Theta(l-l_{r})\beta+\frac{l-a}{b-a}(1-\beta)=\frac{1}{2}
\]
with $\beta=p(H_{1})=1-p(H_{2})$ representing the bias, hence incorporating
the posterior probabilities on the hypotheses derived so far. Setting
the probability to $\frac{1}{2}$ is the maximum entropy requirement,
since only two results are possible. Solving this equation for $l$
leads to the optimal length for the test. Since the hypotheses can
been derived by the general compressor, the bias for the more simple
hypothesis $H_{1}$ will be strong, $\beta\lesssim1$, since the rod
length $l_{r}$ is a variable already present in the representation
and no new length $l_{0}$ has to be introduced. Therefore, the discontinuity
will jump over $\frac{1}{2}$ and the optimal test is going directly
for the simple hypotheses: $l=l_{r}$ with some small $\epsilon$
around it.

In summary, we see that optimal hypothesis tests can be computed by
the maximum entropy principle. Of course, there is no need for mathematical
derivation of the necessary distributions in practice, as we did here.
Instead, the distributions can be bootstrapped, given our generative
hypotheses, leading to approximately optimal tests.

\section{Role of compression and grounding in learning the world's concepts}

How would all that, compression, grounding, help the system understand
the actual world with its complex concepts?

Consider static, black line drawings on a white background as an example
of environment for an AGI system. Eventually, the system shall fulfill
Minsky's call for the ability to talk about the objects in the drawing.
The scene should not be specified in advance and could contain any
everyday scene, like a landscape with houses, cars and trees, or a
room with furniture and various artifacts. How is compression and
grounding useful for reaching such a task?

\subsection{\label{sub:Conceptualizing-objects}Conceptualizing objects and relations
through compression}

First, the system would notice that the $n\times n$ image contains
only black and white points, which reduces the entropy enormously
to $n^{2}$ bits. Further, a good idea is to assume that all points
are white with only few exceptions that constitute the line drawing.
Therefore, it is enough to store that all are white and the positions
of the black points. Each point requires the specification of two
coordinates, taking $2\log_{2}n$ bits. If the number of black points
is $m\ll n^{2}$ then the entropy reduces to $1+2m\log_{2}n\ll n^{2}$
bits (one bit to specify the background color). One may call this
compression and representation step as the discovery of the concept
POINT. Further, if the drawing consists of straight lines, the system
should discover that as well, meaning that the coordinates of some
points can be computed from others given the slope of the line. Essentially,
the number of line end points $l$ is much smaller than the overall
number of black points, which reduces the entropy further to $1+2l\log_{2}n\ll1+2m\log_{2}n$.
This may be called as the discovery of the LINE concept. Consider,
for example a ``house'' drawn as a triangle on top of a square (Fig.\
\ref{fig:Trapezoids}a). It requires only 6 lines to be specified,
hence $l=12$. Assume $n=128$ then, we get an entropy $H=1+2l\log_{2}n=169$
bits, which is quite small compared to a random bit image with $H=n^{2}=16384$
bits. Further, the system could discover that those lines are connected,
i.e.\ the line ends of some lines constitute the same points, which
make a loop. Hence, the drawing can be put together by two closed
POLYGONS. Subsequently, the system may discover that the number of
points between the corners of one of polygons is equal, which gives
birth to the rough concept of LENGTH and SQUARE. The number of corners
then define the concepts TETRAGON and TRIANGLE.

There are several lessons to learn from this procedure. First, the
guiding principle for concept generation is compression. For example,
there is a cascade of subsets: squares $\subset$ rhombs, rectangles
$\subset$ parallelograms $\subset$ trapezoids $\subset$ tetragons
$\subset$ polygons $\subset$ chains of lines $\subset$ set of lines
$\subset$ binary image. As we have seen, every such step from general
to specific constitutes not just a special case, but a \textit{compressible}
special case. A square is not just some specific rectangle, but a
compressible one in the sense that all sides are equal and therefore
less numbers are needed to define it.

Second, each specification step is reached by imposing a constraint
on the previous one. After all, because of the recursivity of the
compression algorithm, every compression step merely compresses the
remaining degrees of freedom of the previous compression step.

Third, ambiguities are resolved by compressibility. After all, there
is no unique partition of the ``house'' into polygons. Nevertheless,
since a square or even rectangle is a quite special polygon the scene
is preferably partitioned into the square and the remaining isosceles
triangle (see ch.\ \ref{sub:Interpretation-rivalry} on interpretation
rivalry).

Finally, concept learning is driven not just by compression, but also
by the stimuli that occur preferably in the world. After all, from
the point of view of compressibility one could attach the ``roof''
just as well on the side of the building instead of its top. Hence,
the world biases concept learning towards actually occurring cases.
This becomes especially important when the number of possible objects
increases exponentially with the number of elements that an object
consists of.

Beyond the representation of objects, spatial relations can also be
derived. For example, the concept of a DISTANCE between objects could
be conceptualized as the length of an imagined line between them.
Further, the concepts ABOVE, BELOW, LEFT, RIGHT are low complexity
conditions on the x- and y-coordinates.

Of course, this brief discussion does not demonstrate the viability
of the approach, nor does it show that all concepts can be derived
this way, including abstract ones. However, it highlights the important
role, compression might play in the generation of concepts about the
world.

\subsection{Grounded knowledge bases}

Despite all advantages of grounded reasoning and representations,
the tight connection to the present input leads to mere fleeting representations
immediately forgotten after the relevant input disappears. This begs
the question about permanent knowledge storage in a way consistent
with the present ideas.

A tentative and admittedly incomplete idea is to store knowledge about
objects in the form of \textit{typical templates}. A template is an
image of an object that is fully stored in the system memory. Storing
complete images is necessary for grounded reasoning since the system
has to preserve the ability to reason about imagined objects at arbitrary
granularity. After all, for sufficiently complex objects such as a
Mercedes, it is doubtful that the system is or should be able to store
a complete feature basis for it. If the basis is not complete, not
all details of the object can be restored by sampling from the basis
onto the mental stage. Therefore, not only are those details excluded
from further reasoning, but basis transformations for resourceful
thinking are impaired as well.

\begin{figure}
\begin{centering}
\includegraphics[width=0.6\textwidth]{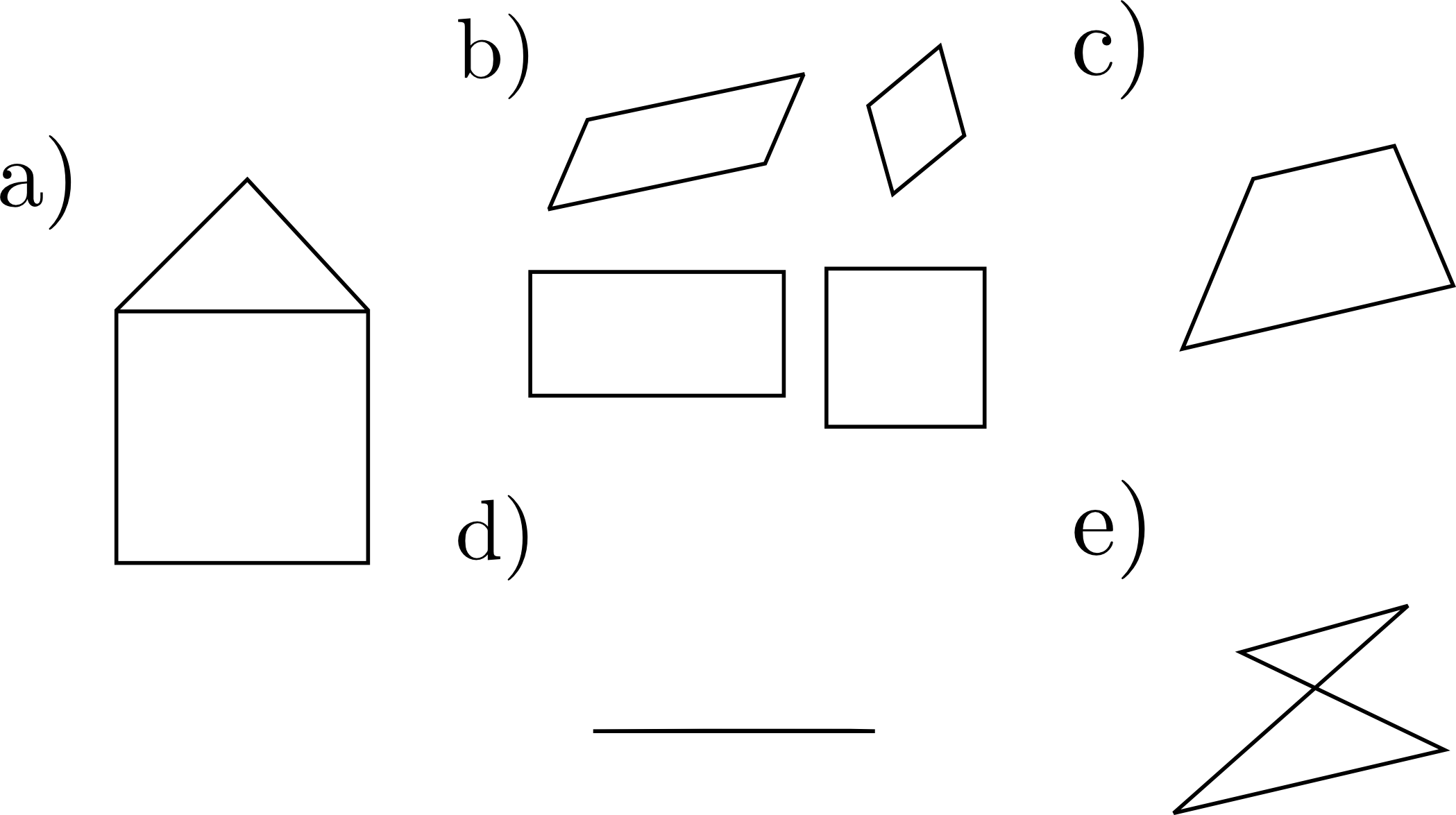}
\par\end{centering}

\caption{A ``house'' (a). Special (b), typical (c), degenerate (d) and ambiguous
trapezoids (e).\label{fig:Trapezoids}}

\end{figure}

However, the template should not contain too much detail -- it should
be typical for the object. A typical template of a concept is an image
of an instance of that concept in such a way that the system's feature
extraction and reasoning processes are able to recognize that concept
as quickly and as unambiguously as possible. For example, consider
different images of a trapezoid. Fig.\ \ref{fig:Trapezoids}b shows
atypical trapezoids since all of them seem to represent more regularities
than are meant to imply. Parallelograms have parallel sides, rhombs
have equal lengths, rectangles right angles and squares both. Therefore,
a typical image of a trapezoid in Fig.\ \ref{fig:Trapezoids}c is
more suitable to convey the concept. Conversely, non-trapezoids such
as general tetragons are not suitable. Moreover, the trapezoid should
not be degenerate, e.g.\ when the distance between the parallel lines
is zero (Fig.\ \ref{fig:Trapezoids}d), nor should the sides cross,
since this would allow the interpretation of two triangles touching
each other at a corner (Fig.\ \ref{fig:Trapezoids}e). Thus, the
a typical template contains exactly the right regularities and properties
in order to conclude the intended concept. In such a way, typical
templates are optimized for storing and transmitting information.

\subsection{Grounded commonsense reasoning}

Beyond the generation of concepts, the system has to be able to reason
about them and answer queries about them correctly. Consider again
the ``house'' in Fig.\ \ref{fig:Trapezoids}a. Given parametrized
representations, the only information to be provided in order to define
the scene is the position and side length of square and one point
defining the roof top. This information is enough to generate a particular
``house'' on the board. After all, compression means that a lot
of information is generated, ``unpacked'', from a small amount of
it.

Consider now the amount of queries that can be answered. What is the
base length of the ``roof'' triangle? Since the concepts of a square
and triangle are activated and the scene is generated, the length
of the base can be read off from the scene. If a perpendicular is
dropped from the top of the house, it lands in the middle of the house
floor? The question can be answered affirmatively by simulating the
perpendicular and \textit{measuring} the position where it splits
the floor. Similarly, one can simulate and answer queries about whether
diagonals through the square cross in the middle and that they cut
the right angle of the square in half. Or that the crossing point
of diagonals is exactly below the top of the house.

It is easy to see that the number of possible queries about the image
increases very quickly with the number of involved elements. The crucial
point is that all those queries can be answered by the system without
actually representing or deducing them in any way from the knowledge
base. The adage ``a picture is worth a thousand words'' reflects
the value of the present proposition. The traditional way to reason
about such commonsense problems is to formalize it with logical statements
and answering queries by backward chaining through the knowledge base.
Apart from the drawbacks mentioned in ch.\ \ref{sub:Formal-logic-for},
I conjecture that the size of the knowledge base needed to answer
all such queries in a scene grows much faster with the complexity
of the scene than the number of parameters needed to generate the
scene from compressed representations and thereby answering all such
queries as well. For example, consider spatial relations between $n$
objects. In principle, there can be $n(n-1)$ (asymmetric) relations
between them. Given a particular scene with given object positions,
the relations between them can readily be read off, as spacial relations
are grounded features of the scene. In a formal model of the scene
though, either all $n(n-1)$ relations have to be stored in the knowledge
base, or general rules of symmetry and transitivity and the like have
to be introduced (e.g.\ ``if above(a,b) then below(b,a)'' or ``if
above(a,b) and above(b,c) then above(a,c)''). Things are already
bad enough since the generalizability of those rules is limited and
requires tremendous foresight by the programmer building a full mathematical
description of the world. Even if such a description can be given
such as in Winograd's famous Blocks World \citep{winograd1971procedures},
it is far from clear that a complete set of rules can be provided
and that would be able to answer all queries. Further, it is well
known that the validity of a moderately true rule may dissolve after
a repeated application (e.g.\ a transitive rule along a chain) --
one of the main difficulties that limited the rise of fuzzy logic.
All those problems dissolve when reasoning is grounded since no rules
need to be applied. Instead, the relations of arbitrary objects can
be extracted directly from the scene, while the context-dependent
generalizability of the observations can still be preserved as argued
in ch.\ \ref{sub:Universal-quantification}.

\section*{Conclusion}

\addcontentsline{toc}{section}{Conclusion}

In this paper I have tried to pour my ideas on artificial general
intelligence and on a path toward it into a coherent whole. There
is hardly anything really new to them, except this particular selection
and the hopefully visible line of thought shaping this selection into
an engineering strategy.

Since much is still half-baked, I would like to sketch the next steps
to be done. First, the test for partial progress described in ch.\
\ref{sub:partial-progress} has to be worked out in practice, which
means setting up a Turing machine and testing until what complexity
level current state of the art compression techniques are able to
stay general. For example, the celebrated Lempel-Ziv algorithm will
be likely to fail at compressing a simple non-stationary sequence.
There will be some complexity level at which all current algorithms
will fail thereby setting up the research goal.

Second, it has to be researched how features are to be found in general.
The hope is that one can start out with basic mathematical concepts
(sets, functions, numbers) that turn out to be applicable quite generally.
For example, after the first concepts such as points, lines etc. are
defined, partonomies made up from them could be constructed thereby
defining more complex objects. Hence, I suspect that the search for
``concept primitives'' could end with the set of simple but general
mathematical concepts.

Third, a strategy for recognizing and dealing with boundary problems
has to be worked out. For example, one may establish that adding a
number to the previous sequence entry works well, but breaks down
at the very first entry, since there is no previous one. The system
has to deal with the brittleness of its own generalizations. Interestingly,
dealing with brittleness is similar to dealing with exceptions: one
has to find the truth conditions of observations. It is always the
same problem: a set of hypotheses explaining an observation has to
be set up, seeing an error due to brittleness in this case. Therefore,
the current framework shows the ability to attack the problem of brittleness.

Finally, a demonstrator shall be built that implements my most important
ideas and achieves a level of generality not encountered before.

\bibliographystyle{plainnat}
\bibliography{Paper}

\end{document}